\documentclass[sn-basic]{sn-jnl}

\jyear{2023}%

\theoremstyle{thmstyleone}%
\theoremstyle{thmstyletwo}%

\theoremstyle{thmstylethree}%

\usepackage{natbib}
\usepackage{booktabs}
\usepackage{soul}
\usepackage{amsmath}
\usepackage{amssymb}
\usepackage{makecell}
\usepackage{bbding}
\usepackage{pifont}
\usepackage{bigstrut}
\usepackage{color}
\usepackage{algorithm}
\usepackage{algpseudocode}
\usepackage{multirow}

\begin{document}

\title[Learning Collaborative Knowledge with Multimodal Representation]{Learning Collaborative Knowledge with Multimodal Representation for Polyp Re-Identification}

\author*[1]{\fnm{Suncheng} \sur{Xiang}}\email{xiangsuncheng17@sjtu.edu.cn}

\author[1]{\fnm{Jiale} \sur{Guan}}\email{gjl886scy@sjtu.edu.cn}

\author[2]{\fnm{Shilun} \sur{Cai}}\email{caishilun1988@qq.com}

\author[1]{\fnm{Jiacheng} \sur{Ruan}}\email{jackchenruan@sjtu.edu.cn}

\author[1]{\fnm{Dahong} \sur{Qian}}\email{dahong.qian@sjtu.edu.cn}



\affil[1]{\orgname{Shanghai Jiao Tong University}, \orgaddress{\city{Shanghai}, \postcode{200240}, \country{China}}}

\affil[2]{\orgname{Zhongshan Hospital of Fudan University}, \orgaddress{\city{Shanghai}, \postcode{200032}, \country{China}}}





\abstract{Colonoscopic Polyp Re-Identification aims to match the same polyp from a large gallery with images from different views taken using different cameras, which plays an important role in the prevention and treatment of colorectal cancer in computer-aided diagnosis. However, traditional methods for object ReID directly adopting CNN models trained on the ImageNet dataset usually produce unsatisfactory retrieval performance on colonoscopic datasets due to the large domain gap. Worsely, these solutions typically learn unimodal modal representations on the basis of visual samples, which fails to explore complementary information from other different modalities. To address this challenge, we propose a novel \textbf{D}eep \textbf{M}ultimodal \textbf{C}ollaborative \textbf{L}earning framework named \textbf{DMCL} for polyp re-identification, which can effectively encourage multimodal knowledge collaboration and reinforce generalization capability in medical scenarios. On the basis of it, a dynamic multimodal feature fusion strategy is introduced to leverage the optimized visual-text representations for multimodal fusion via end-to-end training. Experiments on the standard benchmarks show the benefits of the multimodal setting over state-of-the-art unimodal ReID models, especially when combined with the collaborative multimodal fusion strategy. The code is publicly available at \href{https://github.com/JeremyXSC/DMCL}{https://github.com/JeremyXSC/DMCL}.}


\maketitle

\section{Introduction}
\label{sec1}
\begin{figure}[!t]
\centering
\includegraphics[width=0.80\columnwidth]{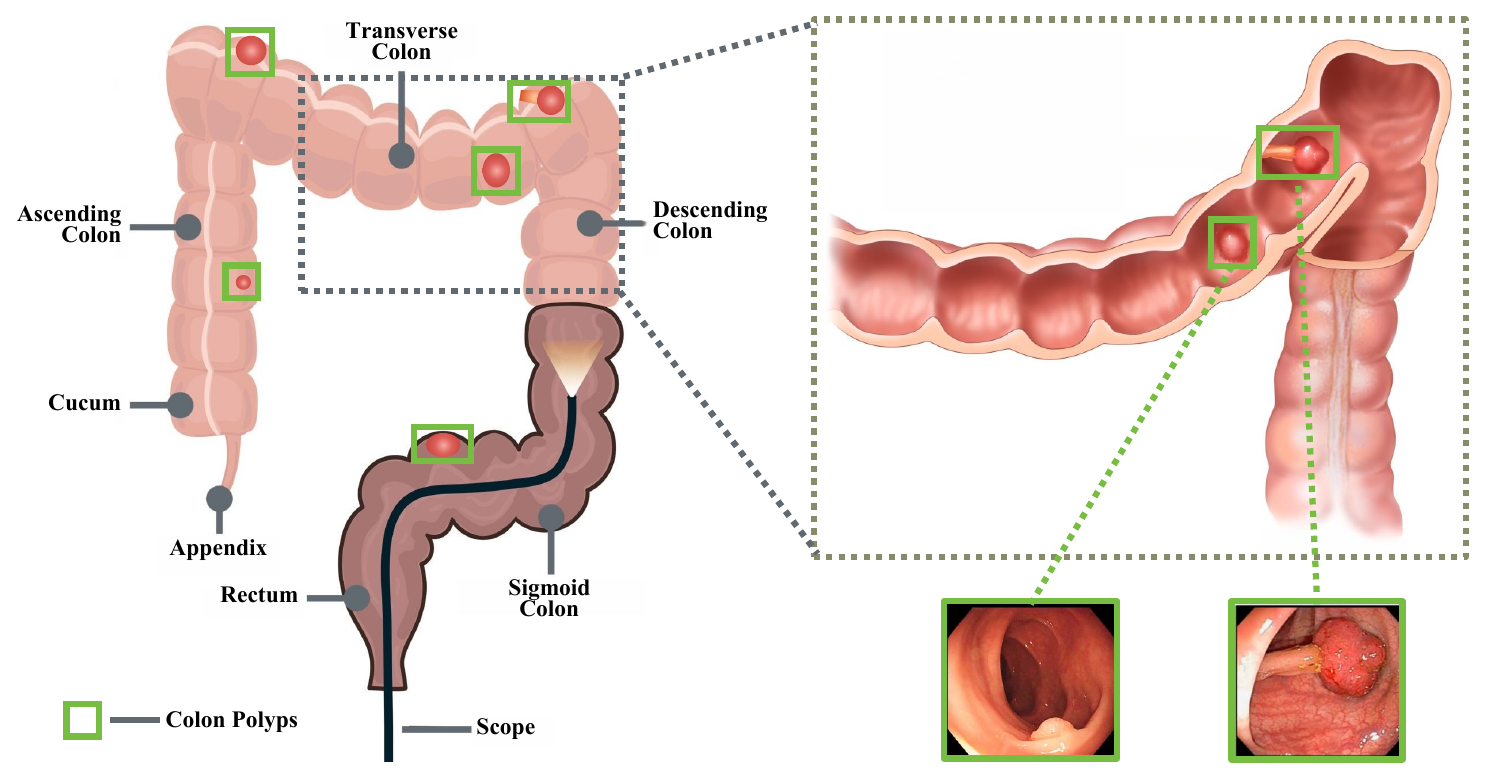}
\caption{The task diagram of colonoscopic polyp re-identification, which matches and correlates polyps appearing across different temporal frames, viewing angles, or clinical examinations. The core technical challenge involves determining whether polyps visualized in distinct images or video sequences represent the same pathological lesion.}
\label{fig-head1}
\end{figure}

Colonoscopic polyp re-identification (Polyp ReID) aims to match a specific polyp in a large gallery with different cameras and locations, which has been studied intensively due to its practical importance in the prevention and treatment of colorectal cancer in the computer-aided diagnosis~\citep{chen2023colo}, it also  provides critical support for clinical applications including longitudinal lesion monitoring, diagnostic decision-making, and therapeutic outcome assessment. Specially, the task diagram of colonoscopic polyp re-identification is illustrated in Fig.~\ref{fig-head1}. With the development of deep convolution neural networks and the availability of video re-identification dataset, video retrieval methods have achieved remarkable performance in a supervised manner~\citep{feng2019spatio,xiang2023rethinking}, where a model is trained and tested on different splits of the same dataset. However, in practice, manually labeling a large diversity of pairwise polyp area data is time-consuming and labor-intensive when directly deploying polyp ReID system to new hospital scenarios. Nevertheless, when compared with the conventional ReID, polyp ReID is confronted with more challenges in some aspects: \textbf{1) from the model perspective:} traditional object ReID methods learn the unimodal representation by greedily “pre-training” several layers of features on the basis of visual samples, while ignore to explore complementary information from different modalities, and \textbf{2) from the data perspective}, polyp ReID  will encounter many challenges such as variation in terms of backgrounds, viewpoint, and illumination, \textit{etc.}, which poses great challenges to the clinical deployment of deep model in real-world scenarios~\citep{xiang2025learning}.

\begin{figure}[!t]
\centering
\includegraphics[width=0.90\columnwidth]{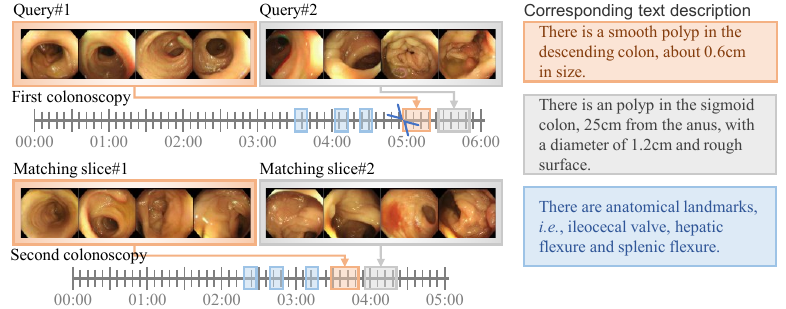}
\caption{The illustration of our multimodal polyp dataset with visual image and its corresponding text description. Specially, given a  query image, the main goal of this work is to learn
a robust polyp re-identification model on the basis of visual-text representation.}
\label{fig-head}
\end{figure}

Based on the aforementioned findings, we novelly propose a \textbf{D}eep \textbf{M}ultimodal \textbf{C}ollaborative \textbf{L}earning framework named \textbf{DMCL} to encourage multimodal knowledge collaboration and reinforce generalization capability in medical scenarios. On the basis of it, a dynamic multimodal training strategy is introduced to leverage the unimodal representations for multimodal fusion via end-to-end training on multimodal tasks, and then improve the performance of our model in an end-to-end manner. The illustration of our multimodal polyp dataset is shown in Fig.~\ref{fig-head}. To the best of our knowledge, this is the first attempt to employ the visual-text feature with collaborative training mechanism for colonoscopic polyp re-identification, from which we have proved that learning representation with multiple-modality can be competitive to methods based on unimodal representation learning. We also hope that our method will shed light on some related research to move forward, especially for multimodal collaborative learning. 

To this end, the major contributions of our work can be summarized as follows:
\begin{itemize}
\item[$\bullet$] We propose a deep multimodal collaborative learning framework (DMCL) to obtain the visual and texture information simultaneously, and then promote the development of multimodal polyp re-identification.

\item[$\bullet$] A simple but effective multimodal feature fusion strategy is introduced to help the model learn more discriminatve information during training and testing stage.

\item[$\bullet$] Comprehensive experiments on benchmarks demonstrate that our system outperforms state-of-the-art methods of polyp ReID by a clear margin, promoting the development of related methods in clinically assisted colorectal polyp recognition.
\end{itemize}
The remainder of this paper is structured as follows. In Section~\ref{sec2}, we give the related works based on hand-crafted based approaches and deep learning based methods in medical area, and then briefly introduce our method. In Section~\ref{sec3}, the details of our multimodal collaborative learning method, as well as the dynamic feature fusion strategy, is presented. Extensive evaluations compared with state-of-the-art methods and comprehensive analyses of the proposed approach are reported in Section~\ref{sec4}. Finally, conclusion of this paper and discussion of future works are presented in Section~\ref{sec5}.

\section{Related Work}
\label{sec2}
\subsection{Hand-crafted based Approaches}
Traditional research works~\citep{prosser2010person,zhao2013person} related to hand-crafted systems for image retrieval task aim to design or learn discriminative representation or pedestrian features. For example, \cite{prosser2010person} propose a reformulation of the person re-identification problem as a learning to rank problem. \cite{zhao2013person} exploit the pairwise salience distribution relationship between pedestrian images, and solve the person re-identification problem by proposing a salience matching strategy. Unfortunately, these handcrafted feature based approaches always fail to produce competitive results on large-scale datasets. The main reason is that these early works are mostly based on heuristic design, and thus they could not learn optimal discriminative features on current large-scale datasets.

\subsection{Deep Learning based Approaches}
Recently, there has been a significant research interest in the design of deep learning based approaches for image or video retrieval~\citep{shao2021temporal,lin2020weakly,wu2018unsupervised}. For example, \cite{shao2021temporal} propose temporal context aggregation which incorporates long-range temporal information for content-based video retrieval. \cite{ma2020vlanet} explore a method for performing video moment retrieval in a weakly-supervised manner, which learns a sharper attention by pruning out spurious candidate proposals. \cite{lin2020weakly} propose a semantic completion network including the proposal generation module to score all candidate proposals in a single pass. As for the self-supervised learning, \cite{wu2018unsupervised} present an unsupervised feature learning approach called instance-wise contrastive learning by maximizing distinction between instances via a non-parametric softmax formulation. However, all above approaches divide the learning process into multiple stages, each requiring independent optimization. On the other hand, despite the tremendous successes achieved by deep learning-based approach, they have been left largely unexplored in terms of semantic feature embedding.

To address these challenges, in our case, we employ the visual-text feature with collaborative learning mechanism for colonoscopic polyp ReID task, which not only improves the polyp ReID performance to a certain extent but also effectively maintains the social logical consistency of the extracted feature, ensuring the reliability and usability of the paradigm in various application scenarios. We hope that our method will shed light on some related researches to move forward, especially for multimodality in medical domain.


\begin{figure}[!t]
\centering
\includegraphics[width=1.0\columnwidth]{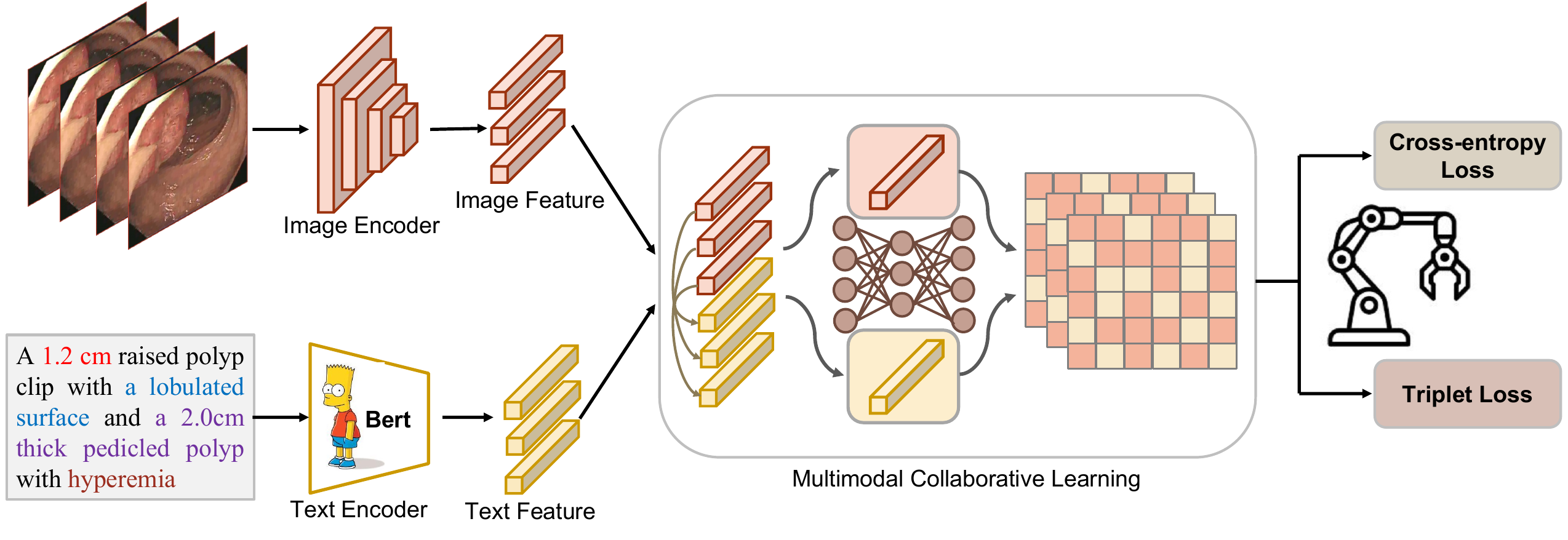}
\caption{The overview of our proposed method  on polyp re-identification task, which contains two main parts: 1) visual feature backbone and 2) textual feature backbone respectively. Specially, visual feature backbone is composed of CNN network, and textual feature backbone is consisted of Transformer architecture named Bert model. In addition, a multimodal fusion strategy is introduced to mine the mutual benefits between visual feature and texture feature, which can further boost the performance of proposed DMCL framework.}
\label{fig-main}
\end{figure}

\section{Our Method}
\label{sec3}

In this section, we firstly give the problem definition of colonoscopic polyp ReID task. Then we introduce the deep multimodal collaborative learning framework (DMCL), as illustrated in Fig.~\ref{fig-main}. Finally, we elaborate more details of our multimodal fusion strategy.

\subsection{Preliminary}
Assuming that we are given a training dataset set $\mathcal{D}$, which contains its own label space $\mathcal{D}=\left\{\left(\boldsymbol{x}_i, y_i\right)\right\}_{i=1}^{N}$ in terms of visual and textual feature, where $N$ is the number of images in the training dataset set $\mathcal{D}$. Each sample $\boldsymbol{x}_i \in \mathcal{X}$ is associated with an identity label $y_i \in \mathcal{Y}=\left\{1,2, \ldots, M\right\}$, where $M$ is the number of identities in the training dataset set $D$, along with the corresponding text description $\boldsymbol{x}_t$ of colonoscopic polyp dataset. The main goal of this work is to learn a robust polyp ReID model on the basis of visual-text representation.

\subsection{Our Proposed DMCL network} 
In this work, we design a DMCL framework for polyp re-identification in medical scenarios, which includes a visual feature extractor, a textual feature encoder and dynamic multimodal collaborative learning strategy.

\textbf{Image Encoder.}
Image encoder is adopted as the visual feature extractor in our DMCL framework. To be more specific, we adopt the ResNet-50~\citep{he2016deep} as the backbone for the image encoder module. Benefit from the merit of residual connection design, ResNet-50 network can effectively address the problem of gradient vanishing, enabling the model to deeply learn image features. On the other hand, our image encoder can extract rich details and global structures to form efficient feature encoding, which exhibits powerful performance and generalization ability in feature extraction. To be more specific, given a 224 $\times$ 224 $\times$ 3-sized polyp image, the main goal of the image encoder is to encode the input into a $I^{1 \times 2048} $-sized vector for subsequent feature fusion and processing, which provides a strong foundation for downstream polyp ReID task. 

\textbf{Text Encoder.}
As for the text encoder module, we employ the BERT model~\citep{devlin2018bert} to encode the text description of corresponding polyp. In essence, BERT model is a pre-trained language model based on the Transformer architecture, which learns language representations through masked language modeling and next sentence prediction tasks on a large amount of unlabeled text. In this work, given a text description of the polyp with $n$ tokens, BERT will encode it into a $n$ $\times$ 768-dimensions vector matrix for subsequent feature fusion and processing, formulated as $[T_{1},  T_{2},  T_{3},  \cdots,  T_{n}]^{1 \times 768}$. The bidirectional encoder design of BERT enables it to capture contextual information of words, generating more accurate text representations.

\textbf{Dynamic Multimodal Collaborative Learning.}
In the general scenarios, there exists an obvious data discrepancy between image and text in our dataset, which inevitably has a negative impact on the model's performance. To overcome this issue, we creatively introduce a multimodal collaborative learning strategy based on the self-attention mechanism, which is a crucial part of our proposed deep multimodal collaborative learning framework. To be more specific, we obtain the image encoding of $I$ after passing through the image encoder, whose size is then dimensionally reduced from the original 2048-dimension to 768-dimension. Similarly, we can also obtain the text encoding through the text encoder, which is denoted as: 
\begin{equation}
T = [T_{1},  T_{2},  T_{3},  \cdots,  T_{n}]
\label{eq3}
\end{equation}
where $T$ denotes the text encoding, and $T_{n}$ represents the number index of character in the polyp's text description. Subsequently, we concatenate the image encoding $I$ and text encoding $T$ to formulate a new encoding of fusion feature, which can be formalized as:  
\begin{equation}
P = [I,  T_{1},  T_{2},  T_{3},  \cdots,  T_{n}]
\label{eq4}
\end{equation}
where $P$ represents the features generated by the multimodal collaborative learning. Remarkably, multimodal fusion component consists of multiple self-attention layers to accomplish the fusion of visual feature and text feature. After passing through three linear layers, these three vectors are then inputted into the self-attention module. The output from the self-attention module is:
\begin{equation}
\text { Output }=\text { Softmax }\left(\frac{\mathbf{Q K}^{\mathrm{T}}}{\sqrt{d}}\right) \mathbf{V}, d=32
\label{eq2}
\end{equation}
where $\mathbf{Q}$, $\mathbf{K}$, $\mathbf{V}$ represents the Query, Key, and Value of self-attention mechanism respectively, and $d$ denotes the dimension of the input vector.
After multimodal collaborative learning, we can obtain the new updated features $P_{\text {c}}$:
\begin{equation}
P_{\text {c}}=\left[I_{c}, T_{c1}, T_{c2}, T_{c3}, \ldots, T_{cn}\right]
\label{eq1}
\end{equation}

Generally speaking, visual feature is of paramount importance for the task of polyp re-identification. Inspired by this, we choose the feature of $I_{c}$ at the corresponding position of the original image feature, which can be adopted as the testing feature of the corresponding polyp in the testing phase for downstream polyp ReID task.
The whole procedure of our collaborative multimodal training system is illustrated in Algorithm~\ref{alg:Framwork}.

\renewcommand{\algorithmicrequire}{\textbf{Input:}}  
\renewcommand{\algorithmicensure}{\textbf{Output:}} 
  \begin{algorithm}[!t]
  \caption{Procedure of our DMCL training system.}
  \label{alg:Framwork}
  \begin{algorithmic}[1]
    \Require \\
      Visual sample $x_{v}$ from  dataset $\mathcal{D}$; \\
      Textual sample $x_{t}$ from  dataset $\mathcal{D}$; \\
      Initialized deep model $\theta$; \\
      Training iterations $n$;
    \Ensure
      Optimized polyp ReID model $\theta_{opt}$; 
    \For{$iter\leq n$}
	\State Random select $x_{v}$ from $\mathcal{D}$;
    \State Random select $x_{t}$ from $\mathcal{D}$;
    \State Extracting visual feature $I$ with $x_{v}$ by $\theta$;
    \State Extracting text feature $T$ with $x_{t}$ by $\theta$; 
    \State Perform multimodal collaborative learning: $P_{\text {c}} \leftarrow I + T$;
    \State Optimizing deep model $\theta$ with $P_{\text {c}}$ $\colon$ $\theta_{opt} \leftarrow \theta$;
    \EndFor
    \State Performing inference with model $\theta_{opt}$;
  \end{algorithmic}
\end{algorithm}

\subsection{Dynamic Network Updating}
In essence, many previous works~\citep{wang2018learning,xiang2024vt} have been found that performing training with multiple losses has great potential to learn a robust and generalizable ReID model. Inspired by this, we adopt a identification loss and triplet loss as the optimization metric for our DMCL model during the training process. Specially, the triplet-loss function aims to reduce the feature distance between similar polyp images while increasing the feature distance between different polyp images. The identification loss function, on the other hand, is to enhance the model's ability to recognize and classify polyp categories.

To be more specific, for a single-label $K$ classification task, the identification loss (cross-entropy loss) is written as,
\begin{equation}\label{eq6}
\mathcal{L}_{ID}=-\frac{1}{\text { $M_{batch}$ }} \sum_{i=1}^{\text { $M_{batch}$}} \sum_{j=1}^K y_{i j} \log \hat{y_{i j}}
\end{equation}
where $M_{batch}$ is the number of labeled training images in a batch, $\hat{y_{i j}}$ is the predicted probability of the input belonging to ground-truth class $y_{i j}$. For a triplet couple \{$x_a$, $x_p$, $x_n$\}, the soft-marginal triplet loss $\mathcal{L}_{triplet}$ can be calculated as:
\begin{equation}\label{eq1}
\mathcal{L}_{Triplet}=\log \left[1+\exp \left(\left\|x_a-x_p\right\|_2^2-\left\|x_a-x_n\right\|_2^2\right)\right]
\end{equation}
where $x_a$, $x_p$ and $x_n$ denote the anchor image, positive sample and negative sample respectively.
As for the triplet selection, we randomly select different polyps from $16$ patients and sample $4$ images or texts from image or text modality for each patient to form the triplet couple, which can greatly help to construct the discriminative embedding for multimodal representation learning.

Finally, the overall objective loss function in a training batch is expressed as:
\begin{equation}
\mathcal{L}_{total}= \mathcal{L}_{Triplet} + \mathcal{L}_{ID}
\label{eq1}
\end{equation}

To this end, our deep multimodal collaborative learning scheme fully considers the information from both images and texts, providing stronger support for clinical colonoscopy polyp examinations and potentially improving model's performance with a clear margin. 

\section{Experiments}
\label{sec4}

\subsection{Datasets and Evaluation Metric}
\label{sec4.1}

We conduct experiments on several large-scale public datasets, which include Colo-Pair~\citep{chen2023colo} for polyp ReID task, and Market-1501~\citep{zheng2015scalable}, DukeMTMC-reID~\citep{zheng2017unlabeled} and CUHK03 dataset~\citep{li2014deepreid} for person ReID tasks. 

To be more specific, \textbf{Colo-Pair}~\citep{chen2023colo} is the first collection of complete paired colonoscopy sequences, which contains 60 videos from 30 patients, with 62 query video clips and the corresponding polyp clips from the second screening manually annotated as positive retrieval clips.  \\
\textbf{Market-1501}~\citep{zheng2015scalable} consists of 32,668 person images of 1,501 identities observed under 6 different camera views. The dataset is split into 12,936 training images of 751 identities and 19,732 testing images of the remaining 750 identities. \\
\textbf{DukeMTMC-reID}~\citep{zheng2017unlabeled} was collected in the winter of Duke University from 8 different cameras, which contains 16,522 images of 702 identities for training, and the remaining images of 702 identities for testing, including 2,228 images as query and 17,661 images as gallery. \\
\textbf{CUHK03}~\citep{li2014deepreid} contains 14.096 images which are collected by the Chinese University of Hong Kong with the images taken from only 2 cameras. Specifically, CUHK03 dataset is divided into 7,368 images of 767 identities as the training set and the remaining 5,328 images of 700 identities as the testing set. In this work, we conduct experiments on the labeled bounding boxes (CUHK03 (Labeled)). 

In our experiments, we follow the standard evaluation protocol~\citep{zheng2015scalable} used in the ReID task and adopt mean Average Precision (mAP) and Cumulative Matching Characteristics (CMC) at Rank-1, Rank-5, and Rank-10 for performance evaluation on downstream ReID task.

\subsection{Implementation details}
\label{sec3.2}
In our experiment, ResNet-50~\citep{he2016deep} is regarded as the backbone with no bells and whistles during visual feature extraction. Following the training procedure in~\cite{xiang2024vt}, we adopt the common methods such as random flipping and random cropping for data augmentation and employ the Adam optimizer with a weight decay co-efficient of $1 \times 10^{-5}$ and $1 \times 10^{-7}$ for parameter optimization. Besides, we adopt the ID loss and triplet loss functions to train the model for 180 epochs, and the cosine distance is also adopted to calculate the similarity of polyp features in the dataset for the task of polyp re-identification. Please note that the text information is also employed during testing phase. In addition, the batch size $M_{batch}$ for training is set to 64. All the experiments are performed on PyTorch framework with one Nvidia GeForce RTX 2080Ti GPU on a server equipped with an Intel Xeon Gold 6130T CPU.

\begin{table}[!t]
\centering
\caption{Performance comparison with state-of-the-art methods on Colo-Pair dataset. \textbf{Bold} indicates the best and \underline{underline} the second best.}
\small
\setlength{\tabcolsep}{0.9mm}{
\begin{tabular}{lccccc}
\toprule
\multirow{2}{*}{Method} & \multirow{2}{*}{Venue} & \multicolumn{4}{c}{Video Retrieval $\uparrow$} \\
\cmidrule{3-6}  &  & mAP & Rank-1 & Rank-5 & Rank-10  \\
\midrule
ViSiL~\citep{kordopatis2019visil} & ICCV & 24.9 & 14.5 & 30.6 & 51.6  \\
CoCLR~\citep{han2020self} & NeurIPS & 16.3 & 6.5 & 22.6 & 33.9  \\
TCA~\citep{shao2021temporal} & WCAV & 27.8 & 16.1 & 35.5 & 53.2  \\
ViT~\citep{caron2021emerging} & CVPR & 20.4 & 9.7 & 30.6 & 43.5  \\
CVRL~\citep{qian2021spatiotemporal} & CVPR & 23.6  & 11.3 & 32.3 & 53.2  \\
CgS$^c$~\citep{kordopatis2022dns} & IJCV & 21.4  & 8.1 & 35.5 & 45.2  \\
FgAttS$^f_A$~\citep{kordopatis2022dns} & IJCV & 23.6  & 9.7 & 40.3 & 50.0  \\
FgBinS$^f_B$~\citep{kordopatis2022dns} & IJCV & 21.2  & 9.7 & 32.3 & 48.4  \\
Colo-SCRL~\citep{chen2023colo} & ICME & 31.5  & 22.6 & 41.9 & 58.1  \\
VT-ReID~\citep{xiang2024vt} & ICASSP & \underline{37.9}  & \underline{23.4} & \underline{44.5} & \underline{60.1}  \\
\midrule
\textbf{DMCL} & \textbf{Ours} & \textbf{46.4} & \textbf{54.3} & \textbf{57.9} & \textbf{60.4} \\
\bottomrule
\end{tabular}}
\label{tab2}
\end{table}

\subsection{Comparison with State-of-the-Arts}
\textbf{Colonoscopic Polyp Re-Identification.}
In this section, we compare our proposed method with the state-of-the-art algorithms,
including: (1) transformer based  (soft attention) models ViT~\citep{caron2021emerging}, Colo-SCRL~\citep{chen2023colo} and VT-ReID~\citep{xiang2024vt}; (2) knowledge distillation-based methods, such as CgS$^c$, FgAttS$^f_A$ and FgBinS$^f_B$~\citep{kordopatis2022dns}; (3) feature level based methods, such as ViSiL~\citep{kordopatis2019visil}, CoCLR~\citep{han2020self}, TCA~\citep{shao2021temporal}, CVRL~\citep{qian2021spatiotemporal}. According to the results in Table~\ref{tab2}, we can easily observe that our method shows the clear performance superiority over other state-of-the-arts with significant Rank-1 and mAP advantages. For instance, when compared to the knowledge distillation-based network FgAttS$^f_A$~\citep{kordopatis2022dns}, our model improves Rank-1 accuracy by \textbf{+44.6\%} (54.3 vs. 9.7). Besides, VT-ReID also surpasses recent transformer based  (soft attention) models ViT~\citep{caron2021emerging}, Colo-SCRL~\citep{chen2023colo} and VT-ReID~\citep{xiang2024vt}. Specially, our method outperforms the second best model VT-ReID~\citep{xiang2024vt} by \textbf{+8.5\%} (46.4 vs. 37.9) and \textbf{+30.9\%} (54.3 vs. 23.4) in terms of mAP and Rank-1 accuracy, respectively. The superiority of our proposed method can be largely contributed to the visual-text representation extracted by our DMCL during multiple collaborative training, as well as the dynamic multimodal feature fusion strategy, which is beneficial to learn a more robust and discriminative model in polyp ReID tasks.

\textbf{Person Re-Identification.}
To further prove the effectiveness of our method on other related object ReID tasks, we also compare our DMCL with existing methods in Table~\ref{tab3}. we can easily observe that our method can achieve the state-of-the-art performance on Market-1501, DukeMTMC-reID and CUHK03 datasets with considerable advantages respectively. For example, our DMCL method can achieve a mAP/Rank-1 performance of 92.1\% and 96.3\% respectively on Market-1501 dataset, leading \textbf{+1.0\%} and \textbf{+0.6\%} improvement of mAP and Rank-1 accuracy when compared to the second best method NFormer~\citep{wang2022nformer} and MGN~\citep{wang2018learning}. In addition, our DMCL method can also obtain the improvement of \textbf{+1.2\%} and \textbf{+1.4\%} in terms of mAP and Rank-1 accuracy on CUHK03 dataset when compared to the VT-ReID~\citep{xiang2024vt}. These experiments strongly demonstrate the priority of our proposed deep multimodal collaborative learning framework.

\subsection{Ablation Studies}

In this experiments, to verify the effectiveness of our deep multimodal collaborate learning framework, we perform ablation study with multimodal feature from both the qualitative and quantitative perspectives.

\textbf{Effectiveness of deep multimodal collaborative learning:}
Firstly, from the quantitative aspect, we evaluate the effectiveness of deep multimodal collaborative learning DMCL framework. As illustrated in Table~\ref{table1}, when adopting our deep multimodal collaborative learning framework, results show that mAP accuracy improves significantly from 25.9\% to 46.4\% on Colo-Pair dataset with visual-text induction. Additionally, similar improvements can also be easily observed in terms of Rank-1, Rank-5 evaluation metrics, leading to +36.8\% and +20.8\% respectively. These results prove that multimodal collaborative learning paradigm has a direct impact on downstream polyp ReID task.

\begin{table}[!t]
  \centering
  \caption{Performance comparison with state-of-the-art methods on Person ReID datasets. \textbf{Bold} indicates the best and \underline{underline} the second best.}
  \small
  \setlength{\tabcolsep}{1.09mm}{
    \begin{tabular}{lcccccc}
    \toprule
    \multirow{2}[4]{*}{Method} & \multicolumn{2}{c}{Market-1501} & \multicolumn{2}{c}{DukeMTMC-reID} & \multicolumn{2}{c}{CUHK03} \\
\cmidrule{2-7}          & mAP   & Rank-1 & mAP   & Rank-1 & mAP   & Rank-1 \\
    \midrule
    PCB~\citep{wang2020surpassing} & 81.6  & 93.8  & 69.2  & 83.3  & 57.5  & 63.7 \\
    Colo-ReID~\citep{xiang2023towards} & 82.1  & 93.3  & 72.3  & 85.9  & 84.6  & 87.0 \\
    MHN~\citep{chen2019mixed}   & 85.0  & 95.1  & 77.2 & 77.3  & 76.5  & 71.7 \\
    ISP~\citep{zhu2020identity}   & 88.6  & 95.3  & 80.0  & 89.6  & 71.4  & 75.2 \\
    CBDB~\citep{tan2021incomplete} & 85.0  & 94.4  & 74.3  & 87.7  & 72.8  & 75.4 \\
    C2F~\citep{zhang2021coarse}   & 87.7  & 94.8  & 74.9  & 87.4  & 84.1  & 81.3 \\
    NFormer~\citep{wang2022nformer} & \underline{91.1}  & 94.7  & \underline{83.5}  & 89.4  & 74.7  & 77.3  \\
    MGN~\citep{wang2018learning}   & 86.9  & \underline{95.7}  & 78.4  & 88.7  & 66.0  & 66.8 \\
    SCSN~\citep{chen2020salience}  & 88.3  & 92.4  & 79.0  & 91.0  & 81.0  & 84.7 \\
    VT-ReID~\citep{xiang2024vt}  & 88.1  & 93.8  & 79.2  & \underline{92.6}  & \underline{85.3}  & \underline{88.3} \\
    \midrule
    \textbf{DMCL} (Ours)  & \textbf{92.1}  & \textbf{96.3}  & \textbf{87.6}  & \textbf{93.5}  & \textbf{86.5}  & \textbf{89.7}  \\
    \bottomrule
    \end{tabular}}%
  \label{tab3}%
\end{table}%

\begin{table}[!t]
  \centering
  \caption{Ablation study of different pre-training settings (\textit{e.g.} \textbf{DMCL w/o Image}, \textbf{DMCL w/o Text} and \textbf{DMCL (Ours)}) from Colo-Pair dataset. Note that the pre-trained model is then fine-tuned on  dataset for downstream polyp ReID task.}
  \small
  \setlength{\tabcolsep}{2.85mm}{
    \begin{tabular}{lccccc}
    \toprule
    \multirow{2}[4]{*}{Pre-training} & \multirow{2}[4]{*}{Text data} & \multirow{2}[4]{*}{Image data} & \multicolumn{3}{c}{Colo-Pair$\uparrow$} \\
\cmidrule{4-6}          &       &       & mAP & Rank-1 & Rank-5    \\
    \midrule
    Baseline & $\times$     & $\times$     & 25.9  & 17.5  & 37.1   \\
    DMCL \textit{w/o} image   & $\checkmark$     & $\times$     & 28.5  & 18.6  & 38.8   \\
    DMCL \textit{w/o} text  & $\times$     & $\checkmark$     & 36.6  & 41.8  & 46.2    \\
    DMCL (Ours)  & $\checkmark$     & $\checkmark$     & 46.4  & 54.3  & 57.9  \\
    \bottomrule
    \end{tabular}}%
  \label{table1}%
\end{table}%

\begin{figure}[!t]
\centering
\includegraphics[width=0.86\columnwidth]{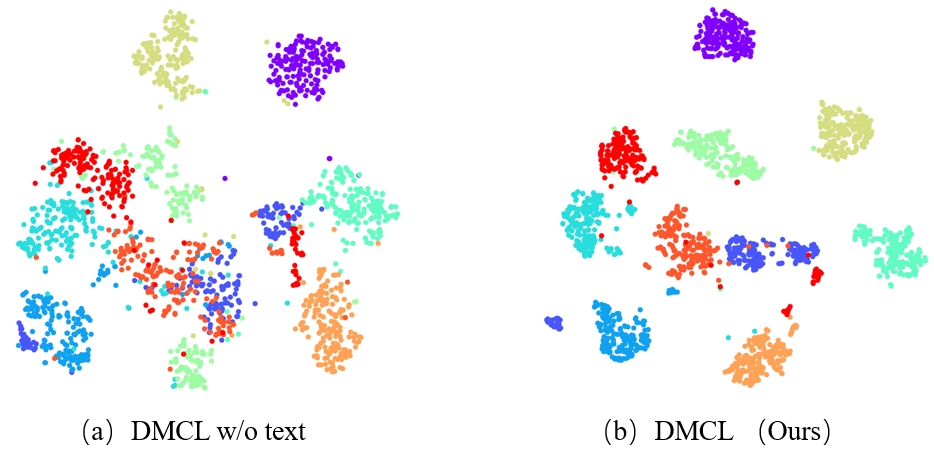}
\caption{The t-SNE visualization of our method with different pre-training settings (\textit{e.g.} (a) \textbf{DMCL w/o Text} and (b) \textbf{DMCL}) on Colo-Pair dataset. Note that points of the same color represent the same class.}
\label{fig-vis2}
\end{figure}

\begin{figure}[!t]
\centering
\includegraphics[width=0.99\columnwidth]{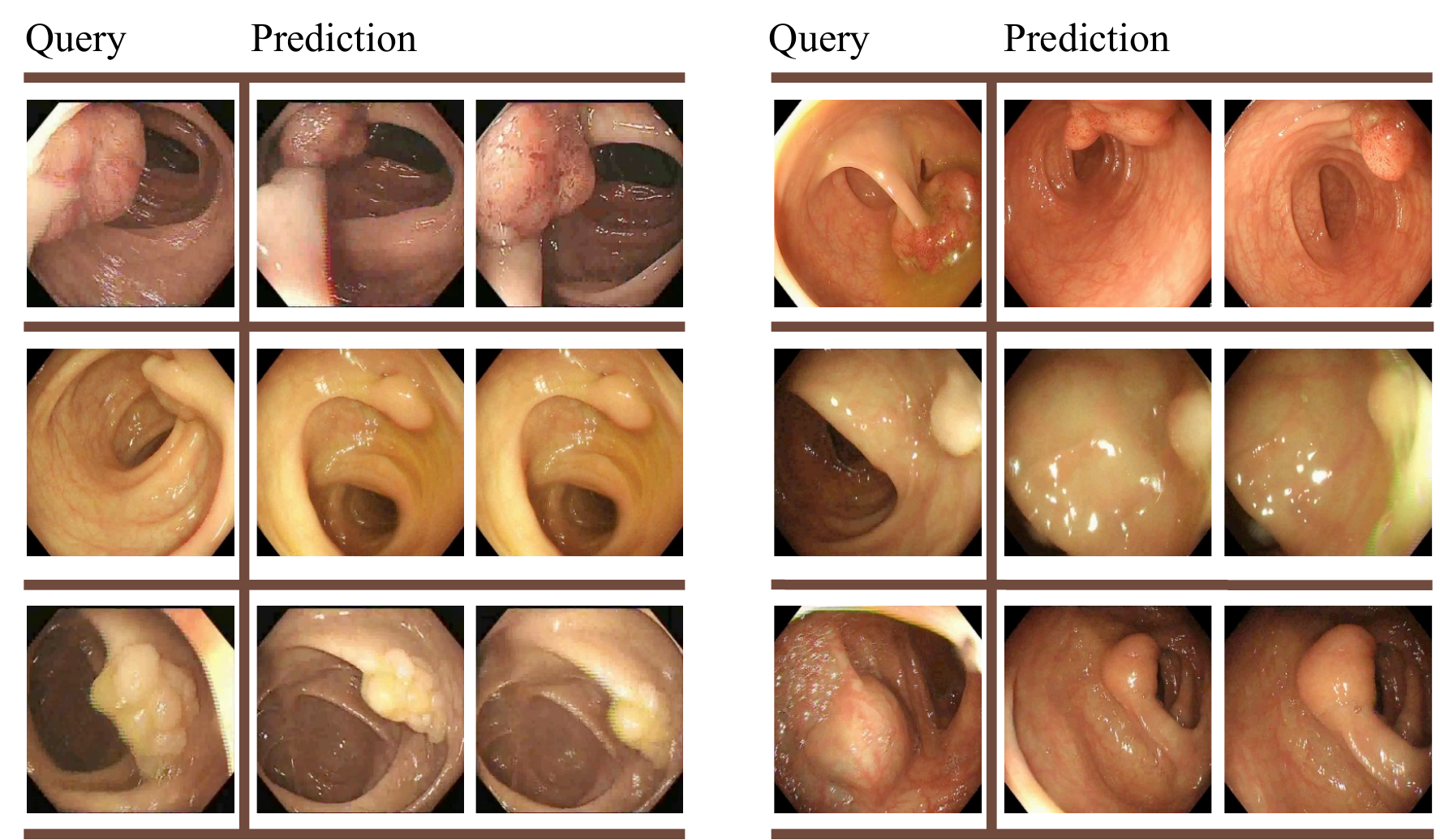}
\caption{Qualitative visualization of ranking results of our proposed approach DMCL with collaborative training mechanism on Colo-Pair dataset.  For each group, the first column shows the query image, while the second and third columns display the inferred results retrieved by our DMCL model. What's more, we can obviously observe that our method shows a great robustness regardness of the deformation or illumination variation of these captured polyps.}
\label{fig-vis}
\end{figure}

Secondly, from the qualitative aspect,  we also give some qualitative results of our proposed deep multimodal collaborative learning framework on multimodal polyp scenarios. For example, according to the t-SNE visualization in  Fig.~\ref{fig-vis2}, it can be obviously observed that our method with multimodal collaborative learning strategy can aggregate the features of the same polyp more tightly, which can significantly enhance the model's robustness and generalziation ability. Furthermore,
Fig.~\ref{fig-vis} also provides some qualitative visualization of the ranking results of proposed approach DMCL with collaborative training mechanism
on Colo-Pair dataset. To be more specific, we can obviously observe that our model attends to relevant image regions or discriminative parts for making polyp predictions, indicating that DMCL can greatly help the model learn more global context information and meaningful visual features with better multimodal understanding, which provides the possibility to discover deeper layer of neural network and more representative features from images automatically, and significantly makes our collaborative training method DMCL model more adaptable in general polyp scenarios. 

\textbf{Effectiveness of dynamic multimodal training strategy:}
In this section, we proceed to evaluate the effectiveness of dynamic multimodal training strategy by testing whether text modality or image modality matters. According to Table~\ref{table1}, our dynamic multimodal training strategy
DMCL with text representation (DMCL \textit{w/o} image) can lead to a significant improvement in Rank-1 of \textbf{+1.1\%} on Colo-Pair dataset when compared with baseline setting.
Furthermore, when adopting image representation, our method (DMCL \textit{w/o} text) can obtain a remarkable performance of 36.6\% in terms of mAP accuracy, leading to a significant improvement of \textbf{+8.1\%} when compared to DMCL \textit{w/o} image.
The effectiveness of the dynamic multimodal training strategy can be largely attributed to that it enhances the discrimination capability of collaborative networks during multimodal representation learning, which is vital for polyp re-identification in general domain where the target supervision is not available.

\begin{figure}[!t]
\centering
\includegraphics[width=1.00\columnwidth]{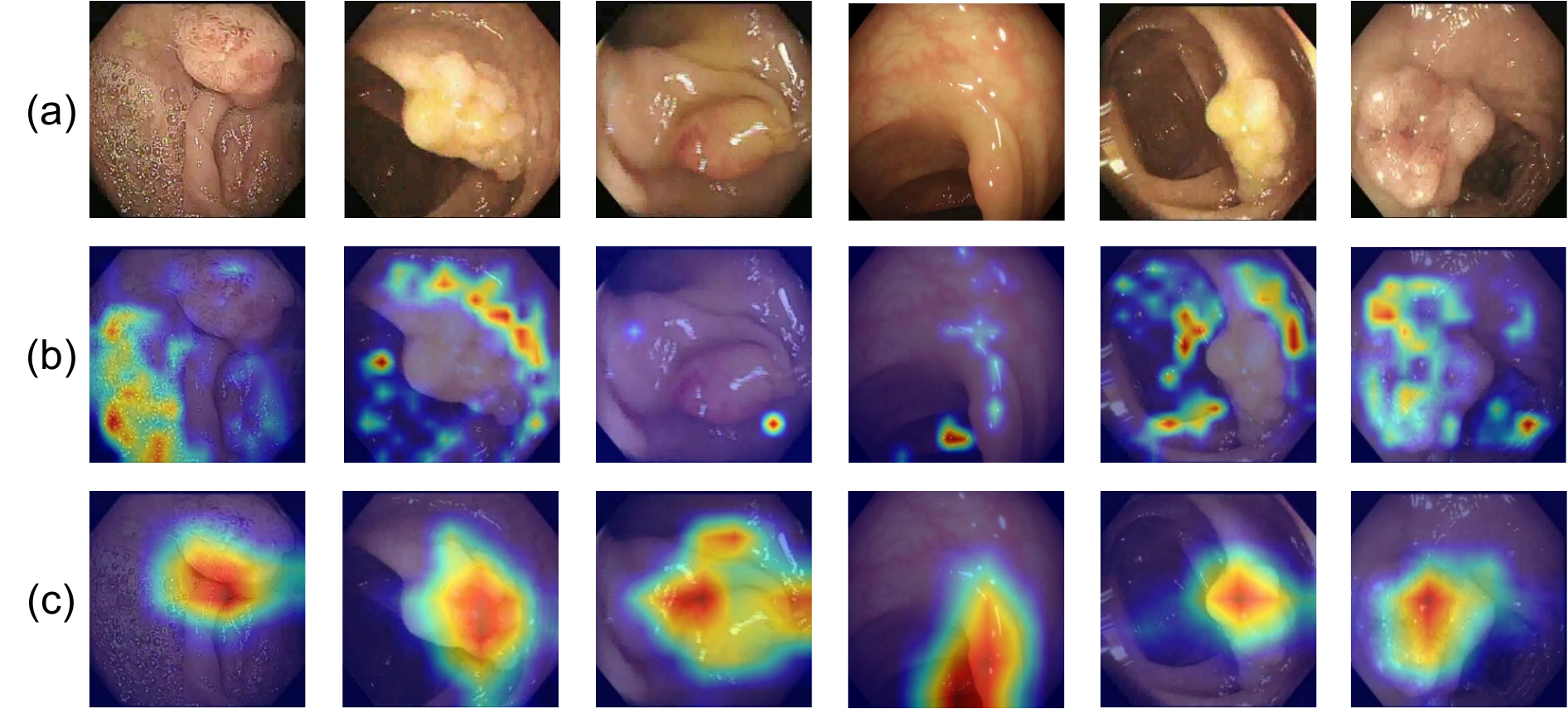}
\caption{Visualization of attention maps with EigenGradCAM: (a) Original images; (b) CNN-based training method without text information; (c) Our deep multimodal collaborative learning method on Colo-Pair dataset. It can be easily observed that semantic pre-training method can capture global context information and more discriminative parts, which are further enhanced in our proposed DMCL method for better performance.}
\label{fig-vis1}
\end{figure}

\textbf{Visualization of Feature Response:}
To further explain why our deep multimodal collaborative learning strategy works, we perform in-depth analysis of feature response in DMCL method, and also show some qualitative examples of EigenGradCAM~\citep{muhammad2020eigen} visualizations in Fig.~\ref{fig-vis1}. In fact, the Grad-CAM is a package with some methods for Explainable AI for computer vision, which provides attributions to both the inputs and the neurons of intermediate layers, so as to make CNN-based models more transparent by producing visual explanations. Specifically, EigenGradCAM decodes the importance of each feature map to a specific class by analyzing the gradients within the last convolutional layer of a CNN, generating a heatmap that highlights the image regions that contribute most significantly to the prediction result. This heatmap visually represents which parts of the image are most influential in the model's final decision. As illustrated in Fig.~\ref{fig-vis1}, compared to the CNN-based training method without text information, we observe that our model attends to relevant image regions or discriminative parts for making decisions, indicating that our DMCL model can effectively explore more global context information and meaningful visual features with better semantic understanding, which significantly make our model more robust to perturbations, such as light-colored areas, disturbance due to its contextual discriminability in visual representation.

\section{Conclusion}
 \label{sec5}
This study further investigates the possibility of applying multimodal collaborative learning for the polyp retrieval task with visual-text dataset, and then proposes a simple but effective multimodal representation learning network named DMCL to improve the performance of polyp Re-ID event. To further enhance the robustness of DMCL model, a dynamic multimodal feature fusion strategy is introduced to leverage the optimized multimodal representations for multimodal fusion via end-to-end training. Comprehensive experiments on standard benchmarks also demonstrate that our method can achieve the state-of-the-art performance in polyp Re-ID task and other image retrieval task. In the future, we will explore the interpretability of this method and apply it to other related computer vision tasks, \textit{e.g.} polyp detection and segmentation.

\bmhead{Acknowledgments}

This work was partially supported by the National Natural Science Foundation of China under Grant No.62301315, Startup Fund for Young Faculty at SJTU (SFYF at SJTU) under Grant No.23X010501967, Shanghai Municipal Health Commission Health Industry Clinical Research Special Project under Grant No.202340010 and 2025 Key Research Initiatives of Yunnan Erhai Lake National Ecosystem Field Observation Station under Grant No.2025ZD03.
The authors would like to thank the anonymous reviewers for their valuable suggestions and constructive criticisms.

\section*{Declarations}

\begin{itemize}
\item \textbf{Funding} \\  This work was partially supported by the National Natural Science Foundation of China under Grant No.62301315, Startup Fund for Young Faculty at SJTU (SFYF at SJTU) under Grant No.23X010501967, Shanghai Municipal Health Commission Health Industry Clinical Research Special Project under Grant No.202340010 and 2025 Key Research Initiatives of Yunnan Erhai Lake National Ecosystem Field Observation Station under Grant No.2025ZD03.
\item \textbf{Conflict of interest} \\  The authors declare that they have no conflict of interest.
\item \textbf{Ethics approval} \\  Not Applicable. The datasets and the work do not contain personal or sensitive information, no ethical issue is concerned.
\item \textbf{Consent to participate} \\  The authors are fine that the work is submitted and published by Machine Learning Journal. There is no human study in this work, so this aspect is not applicable.
\item \textbf{Consent for publication} \\  The authors are fine that the work (including all content, data and images) is published by Machine Learning Journal.
\item \textbf{Availability of data and material} \\  The data used for the experiments in this paper are available online, see Section~\ref{sec4.1} for more details.
\item \textbf{Code availability} \\  The code is publicly available at \href{https://github.com/JeremyXSC/DMCL}{https://github.com/JeremyXSC/DMCL}.
\item \textbf{Authors' contributions} \\  Suncheng Xiang and Jiale Guan contributed conception and design of the study, as well as the experimental process and interpreted model results. Suncheng Xiang obtained funding for the project and provided clinical guidance. Suncheng Xiang drafted the manuscript. All authors contributed to manuscript revision, read and approved the submitted version.
\end{itemize}


\bibliography{sn-bibliography}


\end{document}